\documentclass{article}

\PassOptionsToPackage{numbers, compress}{natbib}

\usepackage{algorithm}
\usepackage{algorithmic}
\usepackage{graphicx}
\usepackage{caption}
\usepackage{subcaption}
\usepackage[final]{neurips_2023}

\usepackage[utf8]{inputenc} 
\usepackage[T1]{fontenc}    
\usepackage{hyperref}       
\usepackage{url}            
\usepackage{booktabs}       
\usepackage{amsfonts}       
\usepackage{nicefrac}       
\usepackage{microtype}      
\usepackage{xcolor}         

\title{O\textsc{bject}C\textsc{omposer}: Consistent Generation of Multiple Objects Without Fine-tuning}

\author{%
  Alec Helbling, Evan Montoya, Duen Horng (Polo) Chau\\
  Georgia Institute of Technology\\
  \texttt{\{alechelbling, emontoya30, polo\}@gatech.edu}
}

\newcommand{\ObjectComposer}{O\textsc{bject}C\textsc{omposer}}

\begin{document}

\maketitle

\begin{abstract}
  Recent text-to-image generative models can generate high-fidelity images from text prompts. However, these models struggle to consistently generate the same objects in different contexts with the same appearance. Consistent object generation is important to many downstream tasks like generating comic book illustrations with consistent characters and setting. Numerous approaches attempt to solve this problem by extending the vocabulary of diffusion models through fine-tuning. However, even lightweight fine-tuning approaches can be prohibitively expensive to run at scale and in real-time. We introduce a method called \ObjectComposer{} for generating compositions of multiple objects that resemble user-specified images.
  Our approach is training-free, leveraging the abilities of preexisting models. We build upon the recent BLIP-Diffusion model, which can generate images of single objects specified by reference images. \ObjectComposer{} enables the consistent generation of compositions containing multiple specific objects simultaneously, all without modifying the weights of the underlying models. 
\end{abstract}

\begin{figure}[htbp]
    \centering
    \includegraphics[width=0.985\textwidth]{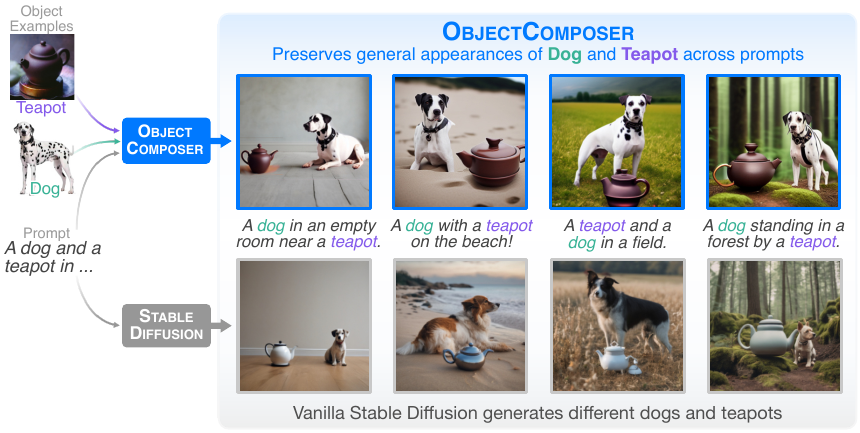}
    \caption{(Top) Given a set of objects $\mathcal{O}$ specified by image and class pairs (in our case a \textit{teapot} and a \textit{dog})  we can generate compositions of those objects in multiple contexts. The general appearance of the \textit{dog} and \textit{teapot} are preserved. (Bottom) We use an out-of-the-box stable diffusion model to generate images of the same captions. For each caption the appearance of the \textit{dog} and \textit{teapot} are different.}
    \label{fig:crown-jewel}
\end{figure}

\begin{figure}
    \centering
\includegraphics[width=\textwidth]{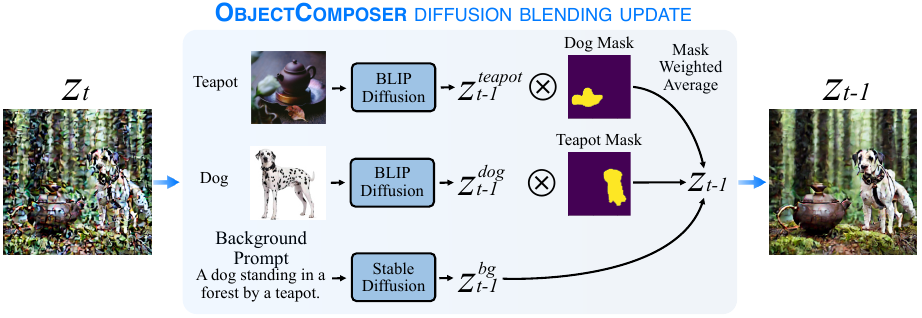}
    \caption{We blend the diffusion processes of several diffusion model's in parallel and compute their average weighted by binary object location masks. }
    \label{fig:blending}
\end{figure}

\vspace{-0.1in}

\paragraph{Motivation \& Contribution}
Recent text-to-image diffusion models \citep{rombach2022highresolution, saharia2022photorealistic, ho2020denoising} have shown the ability to generate high-fidelity images from text prompts. However, these models struggle to generate images of specific objects or subjects (e.g. a specific teapot). The ability to generate images containing multiple specific objects in different contexts is important for many downstream tasks like comic book generation. A common approach to this problem is to fine-tune a diffusion model on a specific object of interest \cite{ruiz2023dreambooth, gal2022image}, binding it to a text identifier that can be referenced later. However, even lightweight fine-tuning is relatively time and compute-intensive, taking several minutes on commercial GPUs. The authors of BLIP-Diffusion \cite{li2023blipdiffusion} develop an approach to generating images of objects specified by images without fine-tuning. However, their approach does not extend to multiple objects simultaneously, and we observe that BLIP-Diffusion struggles to generate complex scenes compared to a vanilla Stable Diffusion model. 
We contribute \textbf{\ObjectComposer{}}, a method allowing users to generate complex scenes containing \textbf{multiple objects} by providing just a \textbf{single image of each object} and a text prompt describing the scene. To the best of our knowledge, ours is the first approach allowing the generation of image compositions containing multiple objects faithful to reference images by using off-the-shelf models. 
\paragraph{\ObjectComposer{}'s Main Ideas} \ObjectComposer{}  takes as input a text prompt $y \in \mathcal{Y}$ and a set of objects $\mathcal{O} = \{(x_i, c_i)\}^n_{i = 1}$ each represented by an image $x_i \in \mathcal{X}$ and a text class $c_i \in \mathcal{C}$ (e.g. ``a dog"). Using this information \ObjectComposer{} generates a composition that matches the description $y$ and contains the objects $\mathcal{O}$. We accomplish this by blending the diffusion processes of an object generation model $\epsilon_o(z_t, t)$ and a background model $\epsilon_b(z_t, t)$. In particular, we leverage BLIP-Diffusion \cite{li2023blipdiffusion} as our object diffusion model $\epsilon_o(z_t, t)$. Because BLIP-Diffusion can only condition upon a single object at once, we compute a separate score $\epsilon_o(z_t, t)$ for each object. Further, we noted that while BLIP-Diffusion excels at generating variations of subjects it struggles to generate more complex compositions faithful to text prompts (see Figure \ref{fig:blip-diffusion-comparison} in the Appendix). Motivated by this, we leverage a vanilla diffusion model, denoted $\epsilon_b(z_t, t)$, to generate the high-level image composition and background. During each timestep $t$ of diffusion inference, for each object we perform a separate diffusion update $z_{t-1}^i = \epsilon_o^i(z_t, t, x_i, c_i)$ conditioned on each object's respective image and class. We also do this for the background $z_{t-1}^b = \epsilon_b(z_t, t, y)$. We combine these latents into a single latent by taking the average weighted by masks that specify the desired locations of each object in the image. These masks can be user-specified, however we leverage the cross-attention maps of a diffusion model to produce these masks (see Appendix for details). We take the pixel-wise average weighted by our binary object masks to produce a final latent $z_{t-1}$ for the next time step in our diffusion process. This blending approach is similar to that used in \cite{bartal2023multidiffusion}. Our approach is shown in Figure \ref{fig:blending}. 

\paragraph{Results}
Our method is capable of generating images containing user-specified reference objects that are also faithful to text prompts. Figure \ref{fig:crown-jewel} shows several examples generated by our approach. Our method consistently generates images containing objects that resemble reference images, in contrast to Vanilla Stable Diffusion model. A limitation of our zero-shot approach is that at times the appearance of an object can deviate from the reference image. For example, our generation of ``a dog with a teapot on the beach" generates a teapot with a similar color but a different shape than the one in our reference images. However, overall our method demonstrates strong potential to allow for the generation of complex scenes with multiple user-specified objects.

\newpage

\paragraph{Ethical Implications}

Text-to-image generation offers exciting new ways to innovate in the visual arts. However, there are ethical concerns. It is possible to leverage text-to-image models to produce derivative artwork that mimics the style of an artist. Further, the tendency for text-to-image models to generate biased content (e.g., doctors as men, nurses as women) is still a developing area of research.  However, when used responsibly we believe our approach can be used to enhance artistic expression, not undermine it. 
Finally, it is possible that composing objects could be used to create deepfakes or other harmful material of people without their consent. However, we believe that research directions similar to ours help to identify the potential risks of generative AI and create public awareness about them. 

\bibliographystyle{plain}
\bibliography{ref}
\newpage
\section{Appendix}

\subsection{Producing masks with cross attention}

To produce masks that specify the location of each of the objects in our desired image we perform diffusion inversion on a given input image $x$. In our experiments, we generate an image $x$ using a Stable Diffusion model to match our prompt $y$, however this image could in theory come from anywhere. We perform null-text inversion \cite{mokady2022nulltext} to invert the diffusion process on a given image $x$. During the process of inversion we store the intermediate cross attention maps between the text and visual tokens of the diffusion U-Net. By averaging these across time we can produce heatmaps that give a fairly strong indication of the location of objects in our image. We apply Otsu's method \cite{4310076} to produce an adaptive threshold for our averaged attention maps, which can be used to produce a binary mask for each object. Leveraging, cross-attention maps to provide layout information for objects has some precedent in recent literature \cite{hertz2022prompttoprompt}.

In our experiments, we also leverage the latent $z_t$ produced by null text inversion as the initialization of our method's diffusion process. 

\subsection{More Results}

\begin{figure}[h]
    \centering
    \includegraphics[width=\textwidth]{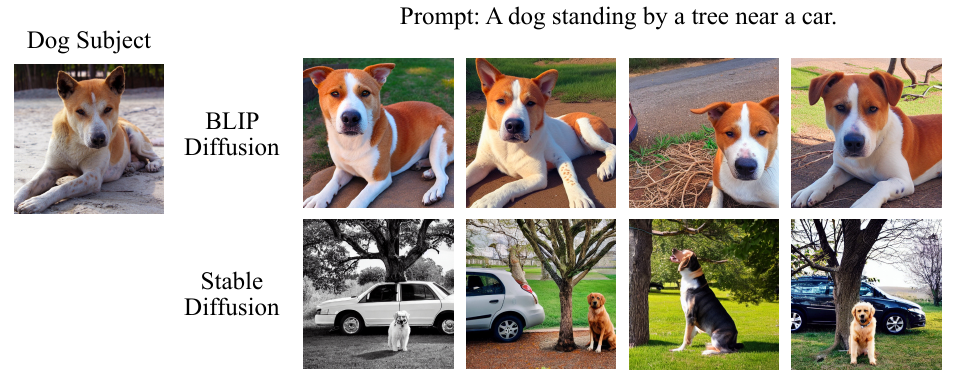}
    \caption{Given a prompt that describes a simple composition like ``a dog standing by a tree near a car" we generate 4 images using BLIP Diffusion given a dog subject. The top row shows 4 images that resemble the dog subject, but fail to otherwise resemble the reference prompt. The bottom row shows the same prompt and 4 corresponding images generated by stable diffusion. They are much more faithful to the given prompt. }
    \label{fig:blip-diffusion-comparison}
\end{figure}


\end{document}